\documentclass{article}

\usepackage[parfill]{parskip}    % Better paragraph spacing
\usepackage[T1]{fontenc}         % Better font encoding
\usepackage[utf8]{inputenc}      % Allow UTF-8 input
\usepackage[margin=1in]{geometry} % Page layout control with specific margins

\usepackage{mathptmx}            % Times font with math support (better than times package)
\usepackage{amsfonts}            % Mathematical symbols
\usepackage{amsmath}             % Advanced math formatting
\usepackage{amssymb}             % Additional mathematical symbols

\usepackage[numbers,sort&compress]{natbib} % Citation management with compressed ranges
\usepackage{authblk}             % Author block formatting
\usepackage{microtype}           % Improves typography and spacing
\usepackage{hyperref}            % Hyperlinks in PDF
\usepackage{cleveref}            % Smart cross-referencing
\usepackage{url}                 % URL formatting
\hypersetup{
    colorlinks=true,
    linkcolor=blue,
    citecolor=blue,
    urlcolor=blue,
    pdftitle={The KL3M Data Project: Copyright-Clean Training Resources for Large Language Models},
    pdfauthor={Michael J Bommarito II, Jillian Bommarito, Daniel Martin Katz},
    pdfkeywords={Large Language Models, Training Data, Copyright, Legal Documents, Government Documents},
    pdfsubject={Machine Learning, Legal Informatics},
    pdfdisplaydoctitle=true,
    bookmarksnumbered=true,
    pdfstartview={FitH},
    pdfpagemode=UseOutlines
}

\usepackage{graphicx}            % Enhanced graphics
\graphicspath{{./figures/}}      % Path to figures
\usepackage{xcolor}              % Color support
\usepackage{float}               % Improved float placement

\usepackage{booktabs}            % Professional tables with \toprule, \midrule, etc.
\usepackage{multirow}            % Multi-row cells in tables
\usepackage{tabularx}            % Flexible width tables
\usepackage{longtable}           % Tables across multiple pages
\usepackage{siunitx}             % Scientific units and number formatting
\usepackage{threeparttable}      % Tables with footnotes
\usepackage{caption}             % Better control of figure/table captions
\usepackage{subcaption}          % For subfigures and subtables

\usepackage{appendix}            % Appendix handling
\usepackage{floatpag}            % Floating pages
\usepackage{listings}            % For code listings
\usepackage{algorithm}           % For algorithm descriptions
\usepackage{algpseudocode}       % Pseudocode formatting
\usepackage{enumitem}            % Better control over lists
\usepackage{todonotes}           % For adding notes during draft editing

\title{
The KL3M Data Project: \\
Copyright-Clean Training Resources \\
for Large Language Models \\
}

\author{\normalsize{Michael J Bommarito II,\textsuperscript{1,2} Jillian Bommarito}\textsuperscript{1} \& Daniel Martin Katz\textsuperscript{1,2,3,4}\\
  \textit{\small \textsuperscript{1} Institute for the Advancement of Legal and Ethical AI (ALEA Institute)}\\
  \textit{\small \textsuperscript{2} CodeX---The Stanford Center for Legal Informatics}\\
  \textit{\small \textsuperscript{3} The Law Lab, Illinois Tech---Chicago Kent College of Law}\\          
  \textit{\small \textsuperscript{4} Center for Legal Technology \& Data Science, Bucerius Law School}\\  
}
  
\date{\normalsize{April 9, 2025}}

\begin{document}

\maketitle

\begin{abstract}
Practically all large language models have been pre-trained on data that is 
subject to global uncertainty related to copyright infringement and breach 
of contract.  This creates potential risk for users and developers due to 
this uncertain legal status. The KL3M Data Project directly confronts this
critical issue by introducing the largest comprehensive training data pipeline
that minimizes risks related to copyright or breach of contract. The foundation
of this project is a corpus of over 132 million documents and
trillions of tokens spanning 16 different sources that have been verified to meet
the strict copyright and licensing protocol detailed herein. We are releasing the
entire pipeline, including 1) the source code to acquire and process these
documents, 2) the original document formats with associated provenance
and metadata, 3) extracted content in a standardized format, 4) pre-tokenized
representations of the documents, and 5) various mid- and post-train resources
such as question-answer, summarization, conversion, drafting, classification,
prediction, and conversational data. All of these resources are freely available
to the public on S3, Hugging Face, and GitHub under CC-BY terms. We are committed
to continuing this project in furtherance of a more ethical, legal, and sustainable
approach to the development and use of AI models.
\end{abstract}
\section{Introduction}

Over the past decade, neural-inspired methods \cite{rumelhart1986learning, hopfield1982neural} 
applied to larger corpora with more parameters
\cite{brown2020language, dubey2024llama, guo2025deepseek} have
driven rapid advancements in language modeling. Large language models (LLMs) 
and their multimodal successors have now solved many challenging real-world tasks
\cite{goh2025gpt, dell2023navigating, katz2024gpt, brin2023comparing}.

This technical success has not, however, come without controversy. Many
critics have raised concerns related to model transparency, environmental 
impact, toxicity, and bias \cite{liesenfeld2024rethinking, longpre2024pretrainer,
bommasani2024foundation}. While these issues
deserve attention, we believe that the most fundamental criticism of LLMs is
that they are trained on data questionably collected, often from the very
individuals and organizations most at risk of being economically displaced by
subsequent use.

Practically all existing LLMs use copyrighted materials obtained without
consent or explicit licensing.  Worse yet, the data has often been obtained from
individuals and organizations who have expressed preferences through licenses or
terms that limit or prohibit their use, modification,
or redistribution.  Despite some efforts to mitigate these issues during training
\cite{minsilo2024} and inference \cite{golatkar2024cpr, ippolito2023preventing,
flemings2024differentially}, leading models continue to reproduce unauthorized
copies \cite{chen2024copybench} or breach contracts, for example, by failing to
provide adequate attribution or incompatibly re-licensing model weights or outputs.

In some situations, "fair use" or "fair dealing" might provide a
 \textit{legal defense} for such practices in the event of litigation
\cite{henderson2023foundation, sag2024fairness}.\footnote{It is worth noting that although some model providers are offering “fair use” as a defense to their data collection practices, many such organizations are also inherently acknowledging
the property rights of creators by entering into licensing deals.} "[M]ore than forty countries
with over one-third of the world's population have fair use or
fair dealing provisions in their copyright laws" \cite{band2013fair}. However,
these principles vary significantly across jurisdictions, and, in general, rely
heavily on fact-specific determinations that must be made by judges and juries. Such
fair use defenses may not generalize and will likely
impose significant costs and uncertainties on the use and development of
this technology.

This legal uncertainty has already materialized in court proceedings. As of April 2025, 
several high-profile cases testing these fair use arguments have either been decided
against the infringing party (\textit{Thomson Reuters v. ROSS Intelligence Inc.}) or have survived early dismissal motions
(\textit{Kadrey v. Meta}, \textit{The New York Times v. Microsoft et al.}).
While statutory relief through Congressional action is oft-cited by pundits,
the reality is that such legislation would likely require not only complex coalition
formation across both parties and chambers, but also critical renegotiation
of international frameworks and treaties like the Berne Convention and the
WIPO Copyright Treaty (WCT). Given the current state of litigation and
geopolitics, legal ambiguity on these topics will likely remain for years.

In light of this reality, we set out on an alternative approach detailed in this
paper - the KL3M Data Project, originally known as the Kelvin Legal Large Language Model Dataset.  At its core, the KL3M Data Project is intended
to support a comprehensive, sustainable data ecosystem for LLM development and use
that builds on positive legal rights and consent.  While the application of AI
will likely remain a socially-contentious issue, we believe that the preferred
path forward should be based on legal and ethical frameworks that do not enshrine
the destruction of property rights through the uncompensated non-consensual redistribution of
intellectual property.  If these systems are to be aligned to embody our beliefs
and values for a better future, they must be built \textit{within}, not outside of,
our shared legal and ethical frameworks.

In this paper, we document and release our first major open milestone for the
KL3M Data Project, summarized in \hyperref[tab:contributions]{Table \ref{tab:contributions}}.
These resources are freely available on S3, Hugging Face, and GitHub under permissive
CC-BY terms that permit general use.

\begin{table}[htbp]
    \centering
    \caption{Primary Contributions of the KL3M Data Project}
    \label{tab:contributions}
    \begin{tabular}{p{0.25\textwidth}p{0.7\textwidth}}
        \toprule
        \textbf{Contribution} & \textbf{Description} \\
        \midrule
        Data Protocol & Our formal protocol for determining whether data can be safely included. \\
        \midrule
        Original Documents & Over 132 million original documents collected under our protocol in their original formats with provenance and metadata. \\
        \midrule
        Extracted Content & Trillions of tokens in standardized document representations as text, Markdown, JSON, XML, HTML, and other formats. \\
        \midrule
        Tokenized Content & Hundreds of modular pre-tokenized data subsets, including a curated 579.8B token snapshot ready for large-scale training. \\
        \midrule
        Mid/Post-Train Resources & 
        \begin{minipage}[t]{\linewidth}
            \small
            $\bullet$ Question-answer pairs (e.g., definitions from the CFR, QA from Agency FAQs)\\
            $\bullet$ Abstractive summarization tasks (e.g., rule abstracts and report summaries)\\
            $\bullet$ Extractive summarization tasks (e.g., keywords from GPO or HTML metadata)\\
            $\bullet$ Classification tasks (e.g., Nature of Suit codes, Agency activity)\\
            $\bullet$ Linear and hierarchical drafting tasks (e.g., patents and contracts)\\
            $\bullet$ Multi-turn conversations (e.g., Congressional hearings, rulemaking)\\
            $\bullet$ Prediction tasks (economic reports, case dockets, legislative histories)\\
        \end{minipage} \\
        \midrule
        Enterprise File Sample & Over 400,000 original PDF, Word, Excel, and PowerPoint documents organized by file type and size. \\
        \midrule
        .gov Database & SQL database with over 3.2 million searchable Federal government websites with complete link structure. \\
        \midrule
        Pipeline Software & The complete source code to acquire and process all data in the collection. \\
        \midrule
        KL3M Data Gallery & Interactive exploration of dataset at \url{https://gallery.kl3m.ai/}. \\
        \bottomrule
    \end{tabular}
\end{table}

The remainder of this paper is organized as follows:

\begin{itemize}
\item \hyperref[sec:legal_problem_space]{Section 2 (Legal Problem Space)} discusses the legal challenges related to copyright and contract law in LLM training data.
\item \hyperref[sec:data_protocol]{Section 3 (KL3M Data Protocol)} outlines our formal protocol for deciding whether content can be included.
\item \hyperref[sec:pipeline_implementation]{Section 4 (Pipeline Implementation)} describes the data collection and processing methodology.
\item \hyperref[sec:data_description]{Section 5 (Dataset Characteristics)} provides an overview of the resulting dataset and artifacts.
\item \hyperref[sec:impact_conclusion]{Section 6 (Impact and Conclusion)} discusses the impact and our future road map.
\end{itemize}

Different readers may wish to navigate this paper according to their specific interests. Technical readers primarily focused on the dataset composition and statistics may proceed directly to \hyperref[sec:data_description]{Section 5 (Dataset Characteristics)} after this introduction. Those interested in data acquisition and processing methodology should focus on \hyperref[sec:pipeline_implementation]{Section 4 (Pipeline Implementation)}. Legal scholars and those concerned with copyright and contract considerations will find \hyperref[sec:legal_problem_space]{Section 2 (Legal Problem Space)} and \hyperref[sec:data_protocol]{Section 3 (KL3M Data Protocol)} most relevant to their interests. For a complete understanding of our approach, methodology, and findings, we recommend proceeding sequentially through all sections.
\section{Legal Problem Space: Copyright and Contract Risks}
\label{sec:legal_problem_space}

Global regulatory frameworks already form complex, dynamic systems that require
significant effort to understand and navigate \cite{vivo2025complex, coupette2021measuring,
bommarito2017measuring}.  Understanding the legal problems that emerge
from the development and use of LLMs therefore unsurprisingly touches on a myriad of legal topics, from export controls and data privacy to
city-specific employment law and state tort claims. From a practical 
perspective,
however, the most material and manageable legal risks emerge from two areas:
copyright and contract law.

Copyright and contract law are traditionally structured to help foster productive and sustainable societies by balancing the interests of creators and rightsholders against the interests of other private and public parties.  In many jurisdictions,
copyright grants creators exclusive rights over original works 
with certain exceptions (e.g., fair use). Contract law, typically through licenses
or terms of use, service, or access, is then leveraged to grant additional rights and impose certain restrictions on a third party's use of works.  Whether and when such contract language is enforceable has been a matter of perennial
dispute in the Internet era, but critically, a contract breach may remain legally
actionable even when copyright law might allow certain uses.  Namely, although there is disagreement regarding how copyright and contract should interact and courts have not universally accepted the prevailing academic view that contract law claims are preempted by copyright law.\footnote{A recent paper \cite{rub2017copyright} surveyed 279 cases and highlighted that the “no-preemption” approach is the prevalent interpretation for copyright preemption in the United States.  However, this perspective is not uniform as noted in \cite{elkin2024back, samuelson2010copyright, samuelson1999intellectual}} This creates a complex environment where millions of works with unique legal statuses and contractual terms
may generate multiple potential legal risks for LLM developers and users.

\subsection{Copyright Risks}
\label{subsec:copyright_risks}

Copyright represents a fundamental societal bargain: creators receive exclusive
rights for a limited period, after which works enter the public domain
\cite{patterson1991nature}. During this exclusivity period, creators control
how their works are used through various licensing frameworks ranging from
highly restrictive to permissive.

The Internet era has enabled unprecedented information sharing through
platforms like \textit{GitHub}, \textit{Getty Images}, and \textit{YouTube},
though disputes over digitized content have persisted since early projects like
Google Books \cite{samuelson2009google} and Napster \cite{rayburn2001after}.
LLMs have dramatically intensified these concerns
\cite{samuelson2023generative}. Most model providers have simply ignored both
website terms of service and explicit content licensing restrictions
\cite{longpre2024large}, prompting creators and organizations to implement
additional protective measures against the harvest of their data for use in A.I. training \cite{longpre2024consent}.

Since the advent of the large-scale public Internet, numerous efforts have
tracked its growth and content
\cite{mcmurdo1995internet, alnoamany2014and}, primarily focused on search
engine development. However, indexing for search differs substantially from
wholesale content collection for training. For example, computational
linguists earlier in the Internet era hesitated to incorporate web content
into corpora due to copyright concerns \cite{ide2002american}.  While some such
materials eventually entered research artifacts \cite{ide2008american}, legal
restrictions limited their adoption and organizations like the Linguistic Data
Consortium (LDC) and European Language Resources Association (ELRA) continued
to follow traditional guidance under copyright and contract law.

Other researchers viewed the Internet primarily as a vast data resource,
collecting various content types
\cite{buck2014n, leskovec2016snap, deng2009imagenet}. \textit{Common Crawl}
\cite{smith2013dirt}, a comprehensive web-scale collection, became the
foundation for many early LLMs. This and subsequent AI training datasets like
C4 \cite{raffel2020exploring}, \textit{The Pile} \cite{gao2020pile}, and \textit{Dolma}
\cite{soldaini-etal-2024-dolma} contain extensive amounts of copyrighted
material.

These collection efforts rely almost entirely on "fair use" justifications,
which require case-by-case evaluation and provide no guaranteed protection.
Even without fair use coverage, alternative approaches like compulsory licensing
systems\footnote{A market-based licensing and royalty system would provide more ethical
treatment than the current practice of seizing creative works without \textit{any}
upfront or ongoing consideration.  Such approaches are explored in recent work
from the U.S. Copyright Office \cite{Jaffe2025}.} could balance creator compensation with continued AI innovation.\footnote{In a letter to the
White House Office of Science and Technology (OSTP), OpenAI argued that
``[A]pplying the fair use doctrine to AI is not only a matter of American
competitiveness — it's a matter of national security... If the PRC's developers
have unfettered access to data and American companies are left without fair use
access, the race for AI is effectively over'' \cite{OpenAI}. While regulatory clarity 
would benefit all stakeholders, it remains unconvincing that requiring fair
compensation to creators would significantly impair innovation rates given the
scale of  private and public funds invested in AI development.}

Scholars working with these datasets have occasionally acknowledged copyright
concerns \cite{schafer2016commoncow, habernal2016c4corpus}, but often with
minimal practical effect. The \textit{Dolma} dataset authors explicitly noted
the changing legal landscape yet still distributed copyrighted materials
because the "sources were publicly available and already being used in
large-scale language model pretraining"
\cite{soldaini-etal-2024-dolma} - circular reasoning that appears to prioritize
technical convenience and leaderboard ranking over addressing legal and ethical considerations.

This reflects the AI ethics field's disproportionate focus on model openness
and alignment at the expense of the legal and moral rights of creators and rights holders. While
transparency and toxicity deserve attention, they are just some but not all of the legal and ethical questions that surround generative AI.  Even datasets claiming to address copyright
issues often fall short. The \textit{Common Corpus} dataset
\cite{arnett2024toxicity} has claimed to contain ``only data that either is uncopyrighted or permissively licensed,'' yet provides no substantive description of its copyright verification process. Unfortunately, even a cursory examination of this dataset reveals a significant quantity of copyrighted or restrictively licensed content that weakens the project's impact.

\subsection{Contract Risks}
\label{subsec:contract_risks}

Beyond copyright, contract law presents another significant but less discussed challenge for LLM training. As noted, copyright and contract law have a complex interface, particularly with respect to the question of the preemption of contract law claims.\cite{elkin2024back}   

Online content is typically provided under
terms of service, terms of use, subscription agreements, or explicit licenses
that grant limited rights to the "user" of a website or service along with
various restrictions. These terms are often lengthy, complex, and difficult to
interpret, leading to the misconception that they are irrelevant or
unenforceable.
However, much of the content on the Internet is indisputably provided to users
as
part of an exchange for value, pecuniary or otherwise, and content creators and
distributors spend substantial time and money drafting and posting these terms.

Much of the content on the Internet is also provided under so-called
"open" or "open source" licenses.  These licenses, which are often described on a scale from permissive to restrictive, operate as "commoditized" contracts that simplify the process of granting and receiving rights and obligations.
Open source has most notably been used in the software space with licenses
like the MIT, Apache, or GPL licenses \cite{metzger2015free}, but similar
legal instruments have also been developed for non-software works, such as
the Creative Commons family of "licenses" or deeds.

In general, both software and non-software contracts may:

\begin{itemize}
  \item require attribution to the original creator(s) or licensor(s), generally
or in a specific manner;
  \item restrict use "intended for or directed towards commercial advantage
or monetary compensation," such as in the Creative Commons Non-Commercial (-NC)
licenses;
  \item restrict the creation of derivative works, such as in the Creative
Commons No Derivatives (-ND) licenses; and
  \item restrict the combination or re-licensing of the work, such as the GNU
General Public License (GPL) or the Creative Commons Share-Alike (-SA)
licenses.
\end{itemize}

Some licenses, like the Affero GPL or CC BY-SA licenses, may even combine two or more of these restrictions.

The Creative Commons family of licenses allows a creator or rightsholder to share data with a recipient while still imposing some set of limitations upon its subsequent use.  When a recipient exceeds the granted rights or fails to meet their obligations, this arguably constitutes a breach of contract. If contractual terms prohibit using content for machine learning, then building an LLM with such content is arguably a contractual violation, separate and apart from any legal risk that arises from copyright.\footnote{Again, there are various perspectives regarding the scope and enforceability of such contract provisions.\cite{rub2017copyright, samuelson1999intellectual, yu2018data}  Some scholars are concerned that excessive reliance on private ordering will undermine certain goals of copyright law. However, in the age of Generative AI, individuals should arguably revisit their personal calculus on this topic. Not only are large AI companies are engaged in the uncompensated and non-consensual usage of the intellectual property of creators but the express goal of many such companies is also to displace the livelihoods of those creators.} 

As documented in \cite{longpre2024consent}, a substantial portion of the content in \textit{Common Crawl} \cite{smith2013dirt} and other corpora is derived from content originally released under terms that prohibit:

\begin{itemize}
\item Automated collection or scraping;
\item Commercial use;
\item Creation of derivative works;
\item Redistribution; and/or
\item Use for machine learning or AI training specifically
\end{itemize}

Such restrictions have proliferated as creators respond to AI development.
Major platforms like Reddit, X (formerly Twitter), and numerous news
organizations have modified their terms specifically to address AI training
\cite{longpre2024consent}. Even content under permissive licenses typically
requires attribution - a requirement that LLM developers rarely satisfy in their model releases or outputs.

Even if courts eventually determine that certain AI training constitutes fair use under copyright law, contract claims based on binding terms may very well remain as independent legal risks. Comprehensive legal compliance requires addressing both copyright and contract concerns simultaneously.

\subsection{Wikipedia: A Case Study in the Complexity of Compliance}
\label{subsec:wikipedia_problem}

Many foundational Internet resources are governed by complex licensing
arrangements that are often overlooked by AI developers.  As the most
notable example, Wikipedia content is frequently included in LLM training
datasets. However, Wikipedia and various other Wikimedia Foundation
projects are governed by the Creative Commons Attribution-ShareAlike (CC
BY-SA) license, which imposes important restrictions on the use of
content.

Originally licensed under the GNU Free Documentation License (GFDL) \cite{roessing2010authorship}, Wikipedia later transitioned to the Creative Commons Attribution-ShareAlike (CC BY-SA) license while maintaining GFDL compatibility for older content. Each individual Wikipedia page is a combined work that manifests licenses from or represented by multiple contributors - terms that even the Wikimedia Foundation itself cannot unilaterally alter. 

The Foundation states clearly in its Terms of Use: "You may
import text that you have found elsewhere or that you have co-authored with
others, but in such case you warrant that the text is available under terms
that are compatible with the CC BY-SA 3.0 license"
\cite{roessing2010authorship}. This creates a complex licensing landscape where
individual contributions may be subject to different requirements.

In response to our direct legal inquiry regarding LLM training on Wikipedia content, the Wikimedia Foundation responded with their interpretation of these compliance requirements \cite{bommarito2024wikimedia}. Their response noted: "We are monitoring what many LLM companies do with Wikimedia data and generally to be upfront, many may not be compliant with the letter of the Creative Commons rules or the spirit of the licenses." When questioned about specific compliance mechanisms, they emphasized that downstream developers must "adhere to the `attribution,' `share-alike,' and other elements of the license." 

Most critically for LLM developers, the Foundation explicitly 
rejected the simplified compliance approaches currently employed by virtually all AI companies: "Providing a \textit{general} notice to customers would not be an adequate solution to compliance [...] [T]he notice would need to be made to everyone the content is shared with, not just customers." This position directly contradicts the practices of commercial LLM developers who include Wikipedia content in their training data.

In the context of building or fine-tuning large language models, it is simple to provide a general attribution notice acknowledging input sources to a given dataset  or model. However, specific attribution to the specific work or works that gave rise to a specific model output is a difficult and expensive, if not impossible, technical challenge.\footnote{It is not clear how any model creator could comply with a specific attribution requirement given current technical limitations. At best, one could construct a system to assign statistical attribution through n-gram matching or other statistical inference, such as as in Appendix C of \cite{brown2020language}.  However, from an attribution perspective, this would both require costly infrastructure and undoubtedly produce false positives and false negatives.} While Wikimedia’s interpretation of the CC BY-SA requirement is not the final word on this important legal question, we did \textit{not} include this content given the risk that it could encumber downstream usage.

Datasets such as \textit{The Pile} \cite{gao2020pile} include Wikipedia content without addressing the fact that
the CC BY-SA license explicitly requires attribution and share-alike provisions.
These datasets and models subsequently trained on them are therefore generally
in breach of contractual obligations under CC BY-SA, and this breach arguably would
persist regardless of any potential fair use defense under copyright law.\footnote{Virtually all model providers have used Wikipedia data in constructing their models. To our knowledge, however, none of them have followed the attribution requirement (BY) as interpreted by the Wikimedia Foundation and none comply with the ShareAlike (SA) requirement.}

The Wikipedia case exemplifies why mere public availability, even under an "open license," does not equate to legal usability. The multiple layers of copyright and contract law, each with its own requirements and restrictions, create a complex web of legal and technical requirements that no major LLM dataset or model provider has adequately addressed. These challenges demand a systematic approach to data collection  that proactively evaluates both copyright status and licensing obligations rather than relying on post-hoc  defenses or ignoring contractual terms.

\section{KL3M Data Protocol}
\label{sec:data_protocol}

The KL3M Data Protocol is our systematic approach to address these legal challenges. Rather than relying on uncertain "fair use" arguments or disregarding contractual obligations, we establish clear, consistent criteria that can be directly evaluated by dataset and model developers - no need for adjudication by judge or jury.

\subsection{Legal Risk Assessment Protocol}
\label{subsec:legal_risk_protocol}

\begin{figure}[h]
\centering
\includegraphics[width=0.9\textwidth]{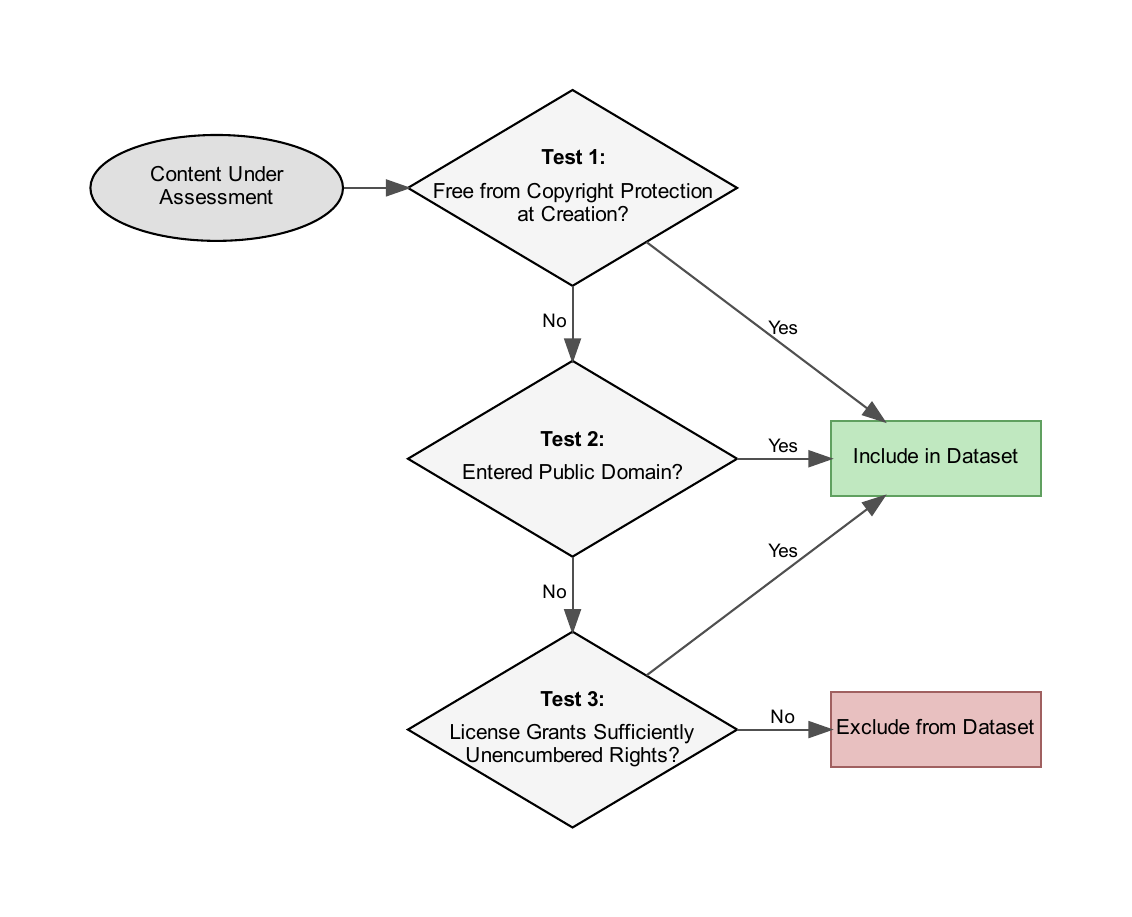}
\caption{The KL3M Data Protocol determines whether to include content through 
three sequential tests that address both copyright and contract risks identified in 
Section~\ref{sec:legal_problem_space}.}
\label{fig:legal_protocol}
\end{figure}

This protocol applies three sequential tests, illustrated in 
Figure~\ref{fig:legal_protocol}, to systematically evaluate both copyright status and contractual terms of potential training content. By implementing this protocol, we create a dataset with substantively reduced legal uncertainty compared to commonly used alternatives.

\subsubsection{Test 1: Free from Copyright Protection at Creation}
\label{subsubsec:test1}

Our first test examines whether content is free from copyright \textit{at the time of its 
creation}. This directly addresses the copyright dimension discussed in 
Section~\ref{subsec:copyright_risks}.

Under 17 U.S.C. § 105, works of the United States government are not generally eligible 
for copyright protection: "Copyright protection under this title is not available for any 
work of the United States Government" \cite{usc17_105}. When federal government employees 
or officers
create documents in their official capacity, these documents automatically enter the 
public domain under this statute.\footnote{With few exceptions as noted in the statute, such as certain academic publications by military academy faculty.}

Similarly, the "government edict doctrine" denies copyright protection to official legal 
materials. This doctrine, which dates back to \textit{Wheaton v. Peters}, 33 U.S. (Pet. 8)
591 (1834) and was most recently addressed in \textit{Georgia v. Public.Resource.Org, 
Inc.}, 590 U.S. 255 (2020) \cite{georgia2020}, ensures that citizens have unrestricted access to the laws governing them. The Copyright Office has stated that this includes "all legislative
enactments, judicial decisions, administrative rulings, public ordinances, or similar 
types of official legal materials" \cite{copyrightcompendium2017}.

Content passing this first test can generally be included in the dataset without further 
copyright or contract evaluation, as it was never subject to these restrictions. Notably, 
many
jurisdictions outside the U.S. do not provide such broad access to government works. For example, the United Kingdom restricts many such works under Crown Copyright or makes them available only under license.

\subsubsection{Test 2: Entered Public Domain}
\label{subsubsec:test2}

If content did have copyright protection at creation, our second test checks whether it 
\textit{has subsequently entered the public domain}. This addresses another aspect of 
copyright risk by identifying materials whose legal status has changed over time.

Content can enter the public domain through:
\begin{itemize}
    \item Expiration of copyright term: Once copyright expires, work automatically enters 
the public domain. For instance, instead of using recent editions of Black's Law 
Dictionary \cite{black2024}, we include the Second Edition from 1910, which contains 
legal definitions that have remained stable for centuries while being free from copyright 
restrictions.
    
    \item Special legal provisions: For example, patent documentation is typically not 
subject to copyright restrictions. The USPTO explicitly notes that "patents are published 
as part of the terms of granting the patent to the inventor" and "the text and drawings 
of a patent are typically not subject to copyright restrictions" \cite{uspto_terms}.
    
    \item Government programs: The U.S. Federal Depository Library Program (FDLP) (44 
U.S.C. § 19) \cite{usc44_19} ensures public access to government information. As
codified in 44 U.S.C. § 1911 \cite{usc44_1911}, "depository libraries shall make
Government publications available for the free use of the general public," and
otherwise-copyrighted content may be entered into the public domain through its
inclusion in the FDLP.
    
    \item Explicit dedication: Content creators may explicitly dedicate their works to 
the public domain through instruments like the Creative Commons Zero (CC0) deed. Several 
notable legal information resources, including CourtListener and the Free Law Project 
\cite{freelaw}, have made such dedications to ensure unrestricted public access to legal 
materials.
\end{itemize}

Content passing either the first or second test can typically be included in the dataset 
without further contract analysis, as no valid copyright protection exists to support most
contractual restrictions.

\subsubsection{Test 3: License Grants Sufficiently Unencumbered Rights}
\label{subsubsec:test3}

The third test specifically addresses the contract risk dimension outlined in 
Section~\ref{subsec:contract_risks}. For content that remains copyright-protected, we 
evaluate whether its license or terms of use \textbf{grant sufficiently unencumbered 
rights} for training language models.

Unlike the binary nature of copyright status, licenses exist on a spectrum from highly 
permissive to highly restrictive. Our analysis considers several key factors:

\begin{itemize}
    \item Whether the license permits commercial use;
    \item Whether the license allows creation of derivative works;
    \item Whether the license imposes "copyleft" or "share-alike" obligations;
    \item Whether attribution requirements can be reasonably satisfied; and
    \item Whether specific prohibitions against machine learning, scraping, or AI
    training exist.
\end{itemize}

This test presents particular challenges with international content and 
jurisdiction-specific licenses. For example, the European Union makes much of its content 
available under relatively permissive terms through Decision 2011/833/EU 
\cite{eu_decision}, which requires only that "the reuser acknowledge the source of the 
[Commission's] documents." In contrast, the United Kingdom employs more restrictive 
frameworks such as the Open Government License (OGL) and the Open Justice License (OJL) 
\cite{ojl}, which impose computational analysis restrictions that would complicate LLM 
training.

This test reflects the complex legal landscape described in 
Section~\ref{subsec:contract_risks}, where online content is typically governed by terms 
of service, terms of use, or standardized licenses that may significantly restrict 
permitted uses.

\subsection{Application to Creative Commons Licenses}
\label{subsec:cc_application}

To illustrate the practical application of this protocol, particularly the third test, we 
analyze how the KL3M Data Project handles the Creative Commons family of licenses. This 
analysis directly connects to our discussion of contract risks in 
Section~\ref{subsec:contract_risks}, where we identified how license terms can create 
legal obligations independent of copyright status.

\begin{itemize}
    \item \textbf{CC0}: \textbf{Always Included} - CC0 represents a complete waiver of 
all copyright and related rights, placing content as close as legally possible to the 
public domain. This license poses no copyright or contract risks.
    
    \item \textbf{CC-BY}: \textbf{Sometimes Included} - The Attribution license only 
requires giving appropriate credit to the creator. We include CC-BY content only where 
attribution requirements can be reasonably satisfied, such as when attribution can be 
provided at scale to a single entity like the European Union under Decision 2011/833/EU 
\cite{eu_decision}.
    
    \item \textbf{CC BY-SA}: \textbf{Always Excluded} - The Share-Alike requirement 
creates "copyleft" obligations that significantly encumber downstream usage. As discussed 
in Section~\ref{subsec:wikipedia_problem}, this requirement would force models trained on 
such content to be released under identical terms, creating incompatibilities with 
standard AI licensing models.
    
    \item \textbf{CC BY-NC}: \textbf{Always Excluded} - Non-Commercial restrictions 
prohibit uses "primarily intended for or directed toward commercial advantage or monetary 
compensation." This directly conflicts with the commercial applications of most language 
models.
    
    \item \textbf{CC BY-ND}: \textbf{Always Excluded} - No Derivatives terms prohibit 
creating "derivative works." As language models inherently learn patterns and generate 
new text based on training data, complying with ND restrictions is technically infeasible.
    
    \item \textbf{Combined restrictions} (e.g., CC BY-SA-NC, CC BY-NC-ND): \textbf{Always 
Excluded} - Licenses combining multiple restrictions create compounded compliance 
challenges and are therefore excluded.
\end{itemize}

This analysis demonstrates how our protocol systematically evaluates both copyright 
status and license terms to determine content eligibility, directly addressing the dual 
legal risks identified in Section~\ref{sec:legal_problem_space}.

\subsection{Application to Real-World Content}
\label{subsec:application}

Our protocol effectively addresses complex real-world content. For instance, Wikipedia 
(discussed in Section~\ref{subsec:wikipedia_problem}) fails our third test because its CC 
BY-SA license creates attribution requirements that cannot be reasonably satisfied and 
would force any model using this content to be released under identical terms, creating 
irreconcilable conflicts with commercial licensing models.

\subsection{Sources and Legal Assessment}
\label{subsec:dataset_sources}

We now provide a comprehensive overview of all sources currently collected in the KL3M Data Project,
including their corresponding legal assessment under the protocol above. Table~\ref{tab:dataset_legal_status} summarizes the 16 primary datasets included in the KL3M collection, organized by their respective legal foundations. For each dataset, we indicate which test(s) it passes and the specific legal basis for inclusion.

\begin{table}[htbp]
\centering
\caption{Legal Status of KL3M Dataset Sources}
\label{tab:dataset_legal_status}
\small
\begin{tabular}{p{4.0cm}p{8.0cm}}
\toprule
\textbf{Dataset} & \textbf{Legal Basis} \\
\midrule
\multicolumn{2}{l}{\textit{\textbf{Test 1:} Free from Copyright Protection at Creation}} 
\\
\midrule
Caselaw Access Project & Government edicts doctrine \newline \textit{Georgia v. 
Public.Resource.Org} \\
Dockets & 17 U.S.C. § 105 \newline Government edicts doctrine \\
Federal Websites & 17 U.S.C. § 105 \newline (U.S. Government works) \\
eCFR & 17 U.S.C. § 105 \newline (Federal regulations) \\
Federal Register & 17 U.S.C. § 105 \newline (Official government journal) \\
GovInfo & 17 U.S.C. § 105 \newline (Government documents) \\
Regulations.gov & 17 U.S.C. § 105 \newline (Regulatory materials) \\
United States Code & 17 U.S.C. § 105 \newline Government edicts doctrine \\
\midrule
\multicolumn{2}{l}{\textit{\textbf{Test 2:} Entered Public Domain}} \\
\midrule
USPTO Patents & Public domain per USPTO Terms \newline 37 CFR 1.71 \\
FDLP & Public domain under 44 U.S.C. § 1911 \\
\midrule
\multicolumn{2}{l}{\textit{\textbf{Tests 1 \& 2:} Both free from copyright and dedicated into the Public Domain}} \\
\midrule
RECAP Archive & 17 U.S.C. § 105 \newline CC0 dedication by Free Law Project \\
RECAP Documents & 17 U.S.C. § 105 \newline CC0 dedication by Free Law Project \\
\midrule
\multicolumn{2}{l}{\textit{\textbf{Test 3:} License Grants Sufficiently Unencumbered 
Rights}} \\
\midrule
SEC EDGAR & Securities law disclosure requirements \\
EU Official Journal & Decision 2011/833/EU \newline (requires attribution only) \\
UK Legislation & Open Government License v3.0 \newline (CC-BY compatible) \\
\bottomrule
\end{tabular}
\end{table}

As shown in Table~\ref{tab:dataset_legal_status}, our datasets fall into three categories 
based on their legal status:

\subsubsection{Test 1: Government Works and Edicts}
\label{subsubsec:test1_materials}

Most datasets pass Test 1, being exempt from copyright at creation. U.S. Government works 
(17 U.S.C. § 105) include federal websites, regulatory documents, and administrative 
publications. Judicial and legislative materials like case law (Caselaw Access Project) 
and statutes are additionally protected by the government edicts doctrine, recently 
affirmed in \textit{Georgia v. Public.Resource.Org}.

\subsubsection{Test 2: Public Domain Materials}
\label{subsubsec:test2_materials}

Several datasets have entered the public domain through:
\begin{itemize}
    \item \textbf{Specific legal provisions}: USPTO patents (per USPTO Terms of Use) and 
Federal Depository Library Program materials (44 U.S.C. § 1911)
    \item \textbf{Explicit dedication}: RECAP court documents through CC0 declarations by 
the Free Law Project
\end{itemize}

\subsubsection{Test 3: Permissively Licensed Materials}
\label{subsubsec:test3_materials}

A smaller set of datasets remains under copyright but permits LLM training through 
unencumbered licensing:
\begin{itemize}
    \item EU Official Journal (Decision 2011/833/EU requiring only attribution)
    \item UK Legislation (Open Government License v3.0, CC-BY compatible)
    \item SEC EDGAR Filings (sufficient rights granted to public through securities law and EDGAR terms)
\end{itemize}

By implementing this three-test protocol consistently across all potential content 
sources, the KL3M Data Project creates a dataset with substantially reduced legal risks
compared to commonly used training resources. Each dataset in our collection has been
systematically evaluated against established legal standards rather than relying on 
uncertain fair use defenses or ignoring contractual obligations altogether.

While no approach can guarantee complete immunity from all potential legal challenges 
in this rapidly evolving area, our protocol establishes a principled foundation that 
respects both copyright status and contractual obligations—providing significantly 
greater 
legal clarity than datasets relying on untested fair use arguments or disregarding 
license terms.
\section{Pipeline Implementation: Data Collection and Processing}
\label{sec:pipeline_implementation}

Establishing a compliance protocol is helpful in theory, but such a protocol only becomes practically useful when executed at scale.  To do so, we carried out an evaluation of hundreds of sources under the protocol in Section~\ref{sec:data_protocol} and selected 16 primary sources to begin with.  We then implemented a general data processing pipeline and source-specific collection automation for the selected sources.  In this section, we describe both the software architecture and selected sources for collection. 

\subsection{From Legal Protocol to Technical Implementation}
\label{subsec:legal_to_technical}

To implement our legal protocol in a technical system, we established a number of key design principles.

First, while we spend significant effort confirming the legal status of sources prior to collection, we believe that it is critical to support the verification of legal status at the document level.  Second, from a preservation and scientific reproducibility perspective, we believe that it is important to allow users to understand and reproduce the process by which original source material is transformed into tokenized training data.  Third, we believe that it is important to support the continuous improvement of this data through quality assessment and future re-processing.  

Based on these principles, we designed a multi-stage approach that, unlike other dataset projects, allows for complete preservation and transparency of training data.

\subsection{Three-Stage Data Flow}
\label{subsec:three_stage_architecture}

\begin{figure}[t]
    \centering
    \includegraphics[width=0.95\textwidth]{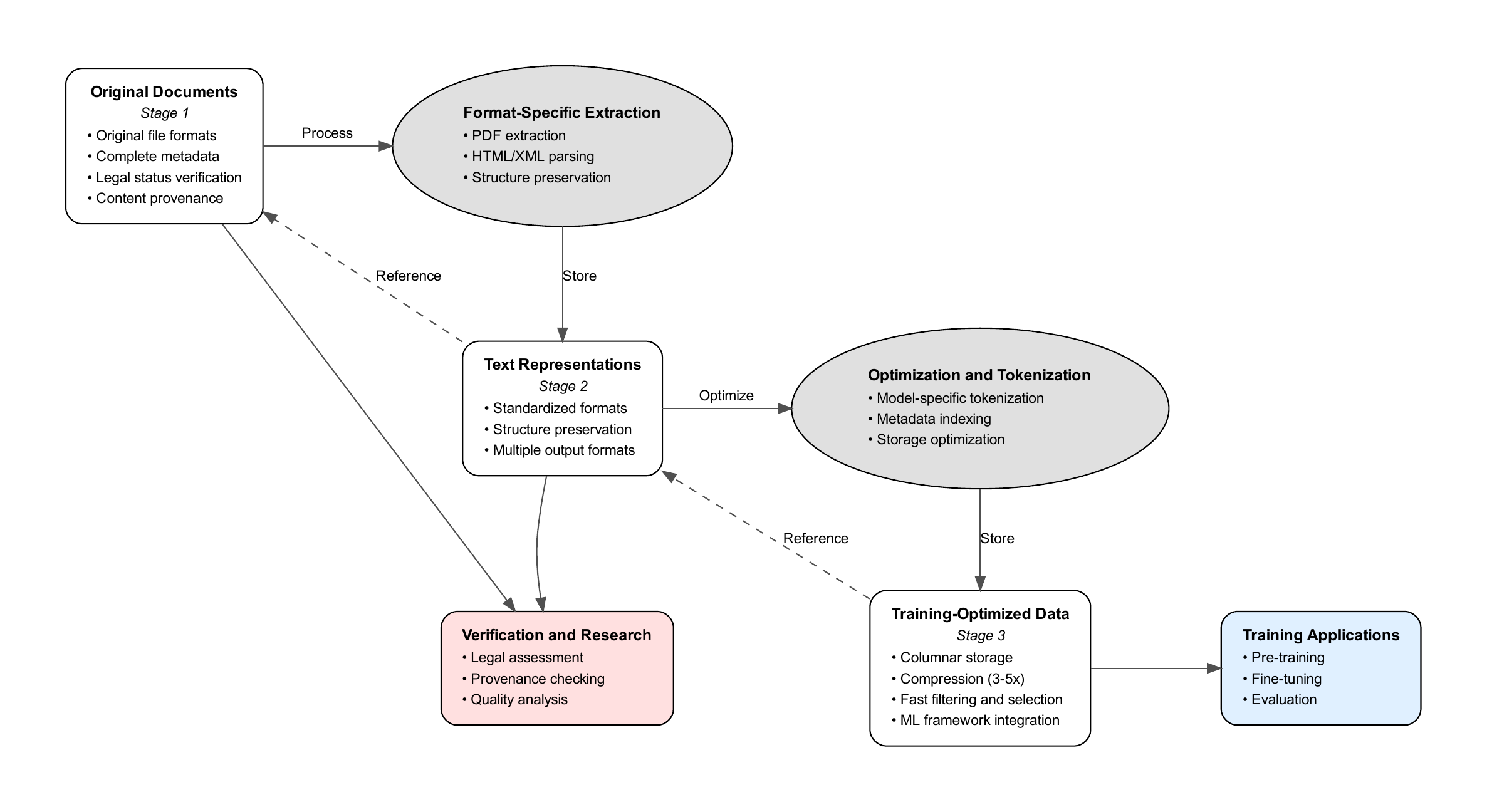}
    \caption{Three-stage data architecture. Documents progress from their original formats with complete metadata to standardized text representations, and finally to storage-optimized formats for model training. Each stage maintains references to previous stages, enabling provenance tracking and future re-processing.}
    \label{fig:document_flow}
\end{figure}

The KL3M data pipeline implements three stages, as visualized in Figure~\ref{fig:document_flow}.  While each stage is designed to serve a different purpose, bidirectional links are maintained through the key structure and metadata to ensure that our pipeline can be tracked as a directed acyclic graph of transformations that preserve all related information.

\subsubsection{Stage 1: Original Documents}
\label{subsubsec:documents_stage}

Original documents are acquired by the \texttt{kl3m\_data.sources} module of the \texttt{kl3m-data} repository in Table \ref{tab:pipeline_components}.  During this stage, raw content in its original format is preserved with comprehensive metadata.  This content is stored in and accessible through the publicly-available S3 bucket \texttt{data.kl3m.ai}, which is located in \texttt{us-east-1} and open under "requester pays" access.\footnote{As a non-profit, we unfortunately cannot afford to pay unrestricted egress fees for the raw data, but users within \texttt{us-east-1} can access this data for free and interested parties may contact us for assistance obtaining the data or coordinating alternative data transfer arrangements.}  

\begin{table}[htbp]
\centering
\small
\begin{tabular}{p{4.5cm}p{9.0cm}}
\toprule
\textbf{Document Field} & \textbf{Description} \\
\midrule
\texttt{content} & Compressed, base64-encoded original file (PDF, HTML, XML, etc.) \\
\texttt{format} & MIME type of the original content (e.g., \texttt{application/pdf}) \\
\texttt{source} & URI or identifier of the original source \\
\texttt{license} & Legal status and license information \\
\texttt{blake2b} & Cryptographic hash for content verification \\
\texttt{id} & Unique document identifier \\
\texttt{dataset\_id} & Source dataset identifier (e.g., \texttt{cap}, \texttt{ecfr}) \\
\texttt{size} & Original content size in bytes \\
\texttt{extra} & Additional metadata from the source API or file \\
\bottomrule
\end{tabular}
\caption{Document stage schema: Core fields preserved for all documents in the collection.}
\label{tab:document_schema}
\end{table}

\begin{figure}[htbp]
\centering
\includegraphics[width=\textwidth]{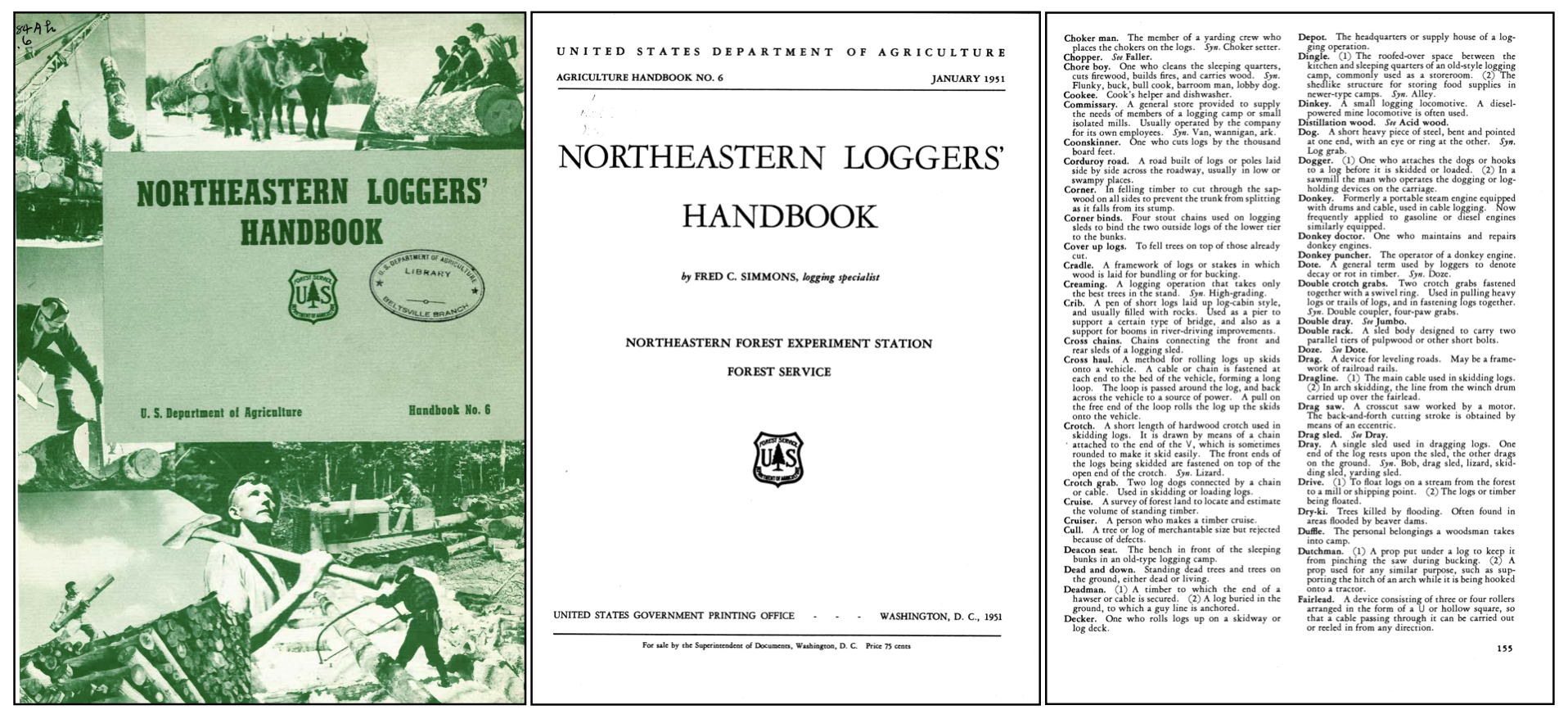}
\caption{The ``Northeastern Loggers' Handbook'' from the Federal Depository Library Program (FDLP) showing the original document preserved in our collection. This document and its associated metadata can be viewed \href{https://gallery.kl3m.ai/document/view?identifier=fdlp/gpo16926/PDF.pdf.json}{here}.}
\label{fig:ne_logging}
\end{figure}

Each of these documents is stored in a structured JSON format under a top-level key that corresponds to the source.  The JSON format is detailed under Table \ref{tab:document_schema}.  To illustrate the schema and richness of preserved metadata, consider the document shown in Figure~\ref{fig:ne_logging}—a 168-page government handbook from 1951 published in the Federal Depository Library Program. The original document, which can be viewed in the KL3M Data Gallery \href{https://gallery.kl3m.ai/document/view?identifier=fdlp/gpo16926/PDF.pdf.json}{here}, includes not only basic bibliographic information, but also specialized fields from the GPO's Catalog of Government Publications.

\begin{itemize}
\item \textbf{Core metadata}: Title, creator, publication date, and description
\item \textbf{Government-specific fields}: SuDoc number (A 1.76:6), Item Number (0003), LC Classification (SD538 .S5)
\item \textbf{Extractable knowledge}: Subject taxonomy (``Logging -- Handbooks, manuals, etc.''), genre classification (``Handbooks and manuals'')
\item \textbf{Provenance chain}: Original URI, print version reference, OCLC number, and CGP PURL
\end{itemize}

This preservation approach ensures that the original source materials remain available for independent verification, legal assessment, and quality control throughout the processing pipeline, even if the original source becomes unavailable.  Source-specific metadata in the \texttt{extra} field also enable specialized analyses and training task creation (e.g., extractive summarization, metadata prediction) while maintaining a consistent core schema.

\subsubsection{Stage 2: Text Representations}
\label{subsubsec:representations_stage}

During Stage 2, original documents are transformed into consistent representations that may be useful for subsequent training or analysis.  The \texttt{kl3m\_data.parsers} module of the \texttt{kl3m-data} repository in Table \ref{tab:pipeline_components} handles determining the content type(s) of a document and attempting to parse the document, including any OCR or content extraction required.  A fallback strategy is used to ensure maximum likelihood of success, and the interested reader may review the source repository and its documentation for full details.  In general, we prefer Markdown content extraction wherever possible, as this format retains valuable structure and formatting like headers, tables, or emphasis relative to plain text.

\textit{Each original document from \textbf{Stage 1} can become one or more documents in this stage.}  For example, in the case of ZIP files retrieved from the Bureau of Reclamation or an S-1 filing by a publicly-traded company, there are often tens or hundreds of embedded documents that are separately extracted.  This one-to-many relationship explains why some tables and figures may show more documents in Stage 2 and Stage 3 than Stage 1. Critically, representation files maintain bidirectional references to their source documents, enabling provenance verification and facilitating reprocessing when extraction techniques improve.

Furthermore, \textit{many files can be transformed into multiple representations.}.  For example, HTML files can be converted to both Markdown and plain text.  When one format cannot be losslessly converted to another, or when multiple parsers are used for quality control, we retain all representations. All resulting representations are stored in the standardized JSON schema with listed under Table \ref{tab:representation_schema}.

\begin{table}[h]
\centering
\small
\begin{tabular}{p{4.5cm}p{9.0cm}}
\toprule
\textbf{Representation Field} & \textbf{Description} \\
\midrule
\texttt{source} & Reference to source dataset \\
\texttt{identifier} & S3 path to original document \\
\texttt{representations} & Map of MIME type to content representation \\
\texttt{representations.content} & Base64-encoded text (typically Markdown) \\
\texttt{representations.tokens} & Pre-tokenized model-specific token IDs \\
\texttt{representations.mime\_type} & Format of the representation (e.g., \texttt{text/markdown}) \\
\texttt{success} & Processing status indicator \\
\texttt{error} & Error message if processing failed \\
\bottomrule
\end{tabular}
\caption{Representation stage schema: Standardized format with tokenization.}
\label{tab:representation_schema}
\end{table}

The Stage 1 and Stage 2 JSON objects for the ``Northeastern Loggers' Handbook'' from Figure \ref{fig:ne_logging} above are stored at the following locations:

\begin{itemize}
    \item Stage 1: \href{s3://data.kl3m.ai/documents/fdlp/gpo16926/PDF.pdf.json}{documents/fdlp/gpo16926/PDF.pdf.json}
    \item Stage 2: \href{s3://data.kl3m.ai/representations/fdlp/gpo16926/PDF.pdf.json}{representations/fdlp/gpo16926/PDF.pdf.json}
\end{itemize}

An abridged sample of the Stage 2 file is provided below:

\begin{verbatim}
{
  "source": "https://permanent.fdlp.gov/",
  "identifier": "s3://data.kl3m.ai/documents/fdlp/gpo16926/PDF.pdf.json",
  "representations": {
    "text/plain": {
      "content": "...",
      ...
    }
  },
  "success": true,
  "error": null
}
\end{verbatim}

\subsubsection{Stage 3: Training-Optimized Data}
\label{subsubsec:parquet_stage}

Since we assume that many consumers of these resources, ourselves included, will be focused on training models, we also prepare training-optimized data formats.  During Stage 3, we convert the Stage 2 representations into Parquet files, which provide an optimized columnar format for efficient training.  Each Parquet file contains references to its source representation file, maintaining the complete provenance chain from training data back to original documents. This format is directly compatible with modern machine learning frameworks like TensorFlow, PyTorch, and JAX, as well as distributed training systems.

The Parquet format is defined via \texttt{pyarrow} in \texttt{kl3m\_data/utils/parquet\_utils.py} as follows using the
\texttt{kl3m-004-128k-cased} tokenizer \cite{bommarito2025tokenizers}, which is a domain-specific BPE model that provides approximately 30-40\% more efficient storage relative to tokenizers like \texttt{gpt-2}.

\begin{verbatim}
DEFAULT_TOKENIZER_NAME = "alea-institute/kl3m-004-128k-cased"
DEFAULT_TOKEN_TYPE = pyarrow.uint32()

schema = pyarrow.schema(
    [
        # source
        pyarrow.field("identifier", pyarrow.string()),
        pyarrow.field(
            "representations",
            pyarrow.map_(pyarrow.string(), pyarrow.list_(DEFAULT_TOKEN_TYPE)),
        ),
    ]
)
\end{verbatim}

\subsubsection{Quality Metrics}
\label{subsubsec:quality_metrics}

To ensure data quality suitable for model training, we developed two quality scoring approaches that evaluates documents based on multiple textual and tokenization characteristics. The first approach, implemented in the \texttt{kl3m\_data.metrics.quality\_metrics} module, uses the metrics listed below to calculate a weighted score representing how far a particular document diverges from control values calculated from high-quality legal sources.  Documents exceeding a certain value can then be excluded from training or flagged for review and reprocessing.

\begin{itemize}
    \item \textbf{Text structure metrics}: Ratios of whitespace, average line length, paragraph length, and document-level organization metrics
    \item \textbf{Character composition metrics}: Ratios of alphanumeric characters, capital letters, punctuation, and non-ASCII characters
    \item \textbf{Token-level metrics}: Type-token ratios, token entropy, character entropy, and repetition rates
    \item \textbf{Format detection metrics}: Identification of formatting artifacts and unusual character sequences
\end{itemize}

A second approach, implemented in \texttt{kl3m\_data.cli.filters} uses a simpler $L^2$ norm between the token-frequency of a document and a control set, filtered to stop words and formatting tokens specially added to the KL3M tokenizers, to again filter or flag documents that are sufficiently different from expected range.

\subsection{Processing Pipeline Implementation}
\label{subsec:processing_pipeline}

The complete processing pipeline is implemented as a modular software stack with four primary open-source components, all released under the MIT license and available on GitHub:

\begin{table}[h]
\centering
\small
\begin{tabular}{p{4.5cm}p{9.0cm}}
\toprule
\textbf{Component} & \textbf{Functionality} \\
\midrule
\href{https://github.com/alea-institute/kl3m-data}{\texttt{kl3m-data}} & Dataset definitions, legal validation, acquisition modules, and pipeline orchestration \\
\href{https://github.com/alea-institute/alea-preprocess}{\texttt{alea-preprocess}} & High-performance document extraction with Rust bindings for CPU-intensive operations \\
\href{https://github.com/alea-institute/alea-dublincore}{\texttt{alea-dublincore}} & Zero-dependency metadata standardization using Dublin Core schema \\
\href{https://github.com/alea-institute/alea-markdown-python}{\texttt{alea-markdown}} & Specialized HTML-to-Markdown conversion with large document support \\
\bottomrule
\end{tabular}
\caption{Core pipeline components: Open-source software stack for document processing.}
\label{tab:pipeline_components}
\end{table}

\subsection{Access and Integration}
\label{subsec:access_integration}

The KL3M Data Project provides multiple access mechanisms with varying levels of abstraction to accommodate diverse research needs:

\begin{table}[h]
\centering
\small
\begin{tabular}{p{3.5cm}p{11.0cm}}
\toprule
\textbf{Access Method} & \textbf{Resource Location and Description} \\
\midrule
\textbf{S3}\footnotemark & \\
Stage 1: Original & \texttt{s3://data.kl3m.ai/documents/} — Complete collection in original formats \\
Stage 2: Representations & \texttt{s3://data.kl3m.ai/representations/} — Standardized text with metadata \\
Stage 3: Parquet & \texttt{s3://data.kl3m.ai/parquet/} — Optimized columnar format for training \\
Enterprise File Sample & \texttt{s3://data.kl3m.ai/raw/} — Word, PDF, etc. documents from .gov domains \\
\midrule
\textbf{Database Access} & \\
SQLite Database & \texttt{s3://data.kl3m.ai/db/dotgov-documents.db} — searchable metadata \\
\midrule
\textbf{Platform Integration} & \\
Hugging Face & \href{https://huggingface.co/alea-institute}{\texttt{alea-institute}} — Pre-tokenized datasets and benchmarks \\
GitHub & \href{https://github.com/alea-institute}{\texttt{github.com/alea-institute}} — Source code and documentation \\
\bottomrule
\end{tabular}
\caption{Access methods for the KL3M resources: Comprehensive options for data and code access.}
\label{tab:access_methods}
\end{table}
\footnotetext{The S3 bucket is configured as a requester-pays bucket due to the substantial data volume. While this means that data transfer costs are borne by the requester, we actively support researchers and interested parties. Please contact the authors for access assistance, alternative data transfer arrangements, or information about public snapshots that do not incur egress fees.}

By providing multiple access strategies, we hope to enable both large-scale model training using optimized formats and detailed examination of specific documents or subsets. Importantly, all access methods maintain the provenance linkages across processing stages via S3 URIs, ensuring that users can always trace from model training data back to original source documents.

\section{KL3M Data Characteristics and Statistics}
\label{sec:data_description}

While we have described this work as a Data \textit{Project}, our hope is that unlike a formal project, our data collection will have no definite end.  Instead, the KL3M Data Project should be viewed as an ongoing operation or "living" dataset.  That said, we appreciate that a description of its current scale and characteristics is required for researchers to understand and evaluate the resources provided. In this section, we examine the key characteristics of the data currently available, focusing on its scale, diversity, and utility for language model training. 

\subsection{Overall Data Scale and Structure}
\label{subsec:dataset_scale}

\begin{figure}[htbp]
\centering
\includegraphics[width=\textwidth]{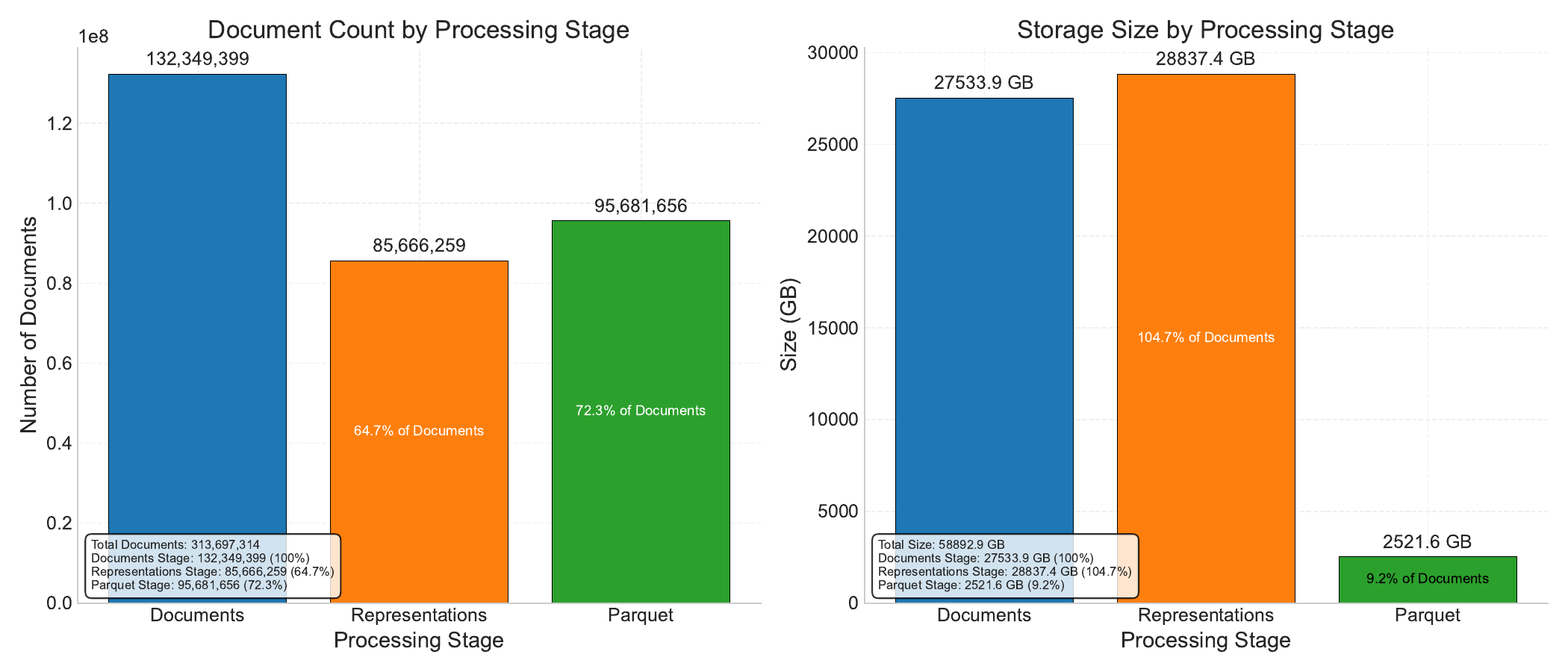}
\caption{Document counts and storage size across the three stages in \texttt{s3://data.kl3m.ai/}}
\label{fig:processing_stages}
\end{figure}

As detailed in Section~\ref{sec:pipeline_implementation}, the primary processing effort is divided into three stages. Figure~\ref{fig:processing_stages} provides a visualization of the relative size, in both S3 objects and bytes, for each
stage as of approximately April 5-6th, 2025.

The current collection includes 132.3 million documents in their original formats, stored as base64 zlib-compressed fields within the JSON format detailed in Section \ref{subsubsec:documents_stage} above; these original documents currently total approximately 28 TB of storage on S3 under the \texttt{s3://data.kl3m.ai/documents/} prefix.  While there are approximately 65-75\% as many documents in Stage 2 and Stage 3, note that 1) we are continuing to process documents based on resource availability and format and 2) each original document may contain one or more embedded documents as detailed in Section \ref{sec:pipeline_implementation} above.

\subsection{Content Diversity and Domain Coverage}
\label{subsec:content_diversity}

Beyond its size, the intellectual diversity of our sources is one of the project's most valuable features. Documents published in our sources cover an impressive range of knowledge, from laws and regulations to scientific research, public commentary, and business strategy.

\begin{figure}[htbp]
\centering
\includegraphics[width=130mm]{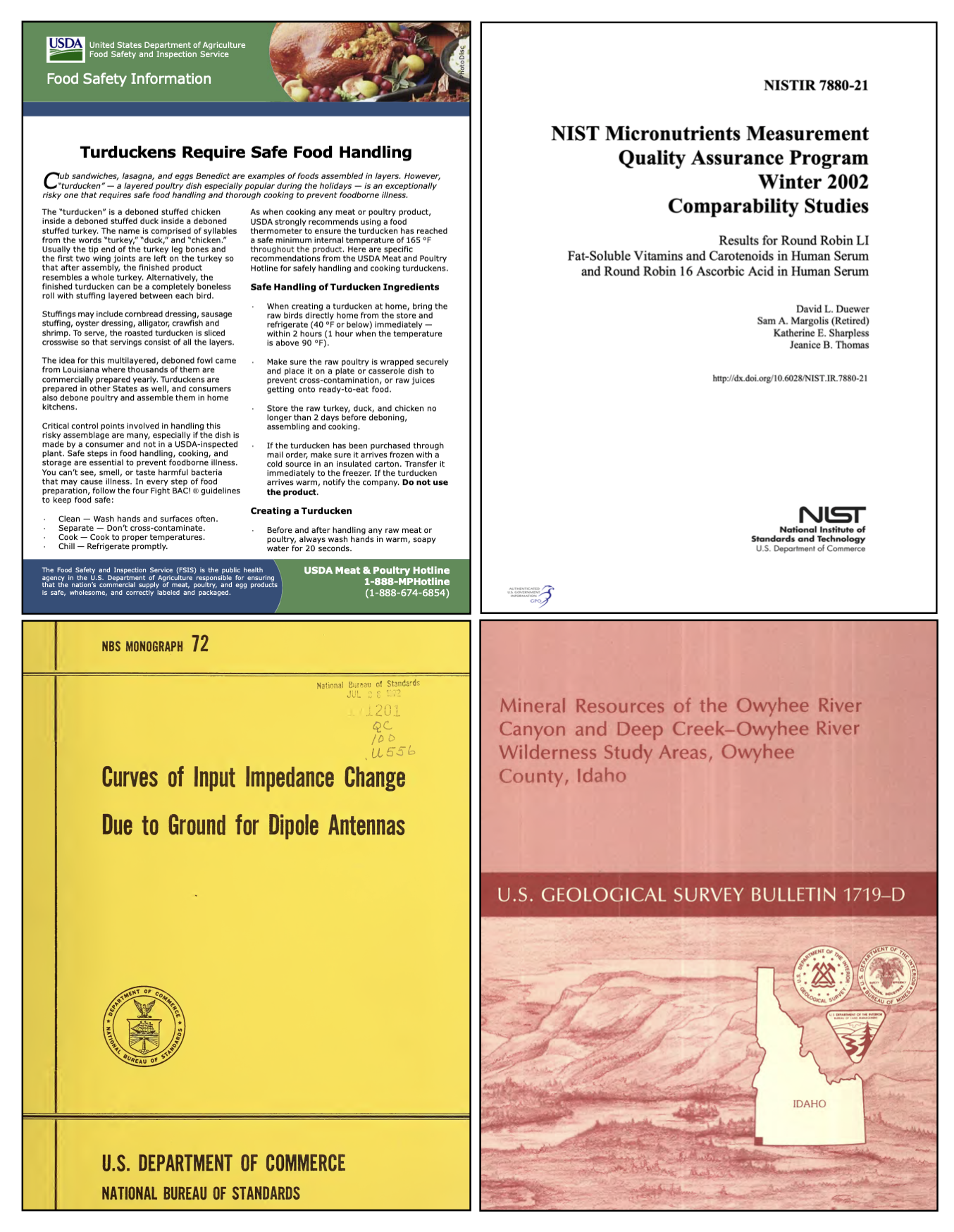}
\caption{\centering{Sample of four documents that illustrate the subject matter diversity. \newline View more at \url{https://gallery.kl3m.ai/}}}
\label{fig:KL3MImages}
\end{figure}

Figure~\ref{fig:KL3MImages} shows four representative examples from the collection: USDA food safety guidelines for turducken preparation, NIST protocols for vitamin testing, Department of Interior geological surveys, and technical engineering analyses from the Department of Commerce. These examples demonstrate the wide range of subjects covered beyond traditional legal documents.

This intellectual breadth provides several distinct advantages for language model training:

\begin{itemize}
    \item \textbf{Domain knowledge depth:} Government documents often represent authoritative, 
fact-checked information that has undergone expert review, providing higher factual reliability 
than many web-scraped alternatives.
    
    \item \textbf{Specialized vocabulary:} Technical documents contain domain-specific terminology 
that helps models develop accurate representations of specialized concepts essential for 
professional applications.
    
    \item \textbf{Procedural knowledge:} Government publications frequently include detailed 
protocols, methodologies, and step-by-step procedures that can enhance a model's ability to provide 
structured guidance.
    
    \item \textbf{Cross-domain connections:} Materials often bridge multiple fields (e.g., where 
science meets regulation or technology meets policy), fostering connections that support 
interdisciplinary reasoning.
\end{itemize}

The KL3M Data Gallery (available at \url{https://gallery.kl3m.ai/}) provides researchers with an interactive tool to explore this diversity more extensively. This web-based exploration tool allows users to browse sample documents across all datasets, view document previews in their original formats, and understand the range of content types available. The Gallery serves as both a research tool for understanding the collection's composition and a practical demonstration of the document preservation approach described in Section~\ref{sec:pipeline_implementation}.

\subsection{Dataset Composition}
\label{subsec:dataset_composition}

Source-specific and aggregate token document counts and estimated token counts currently available are provided in \ref{tab:token_length_stats}.  Notable observations include:

\begin{table}[htbp]
\centering
\caption{KL3M Dataset Document Counts by Processing Stage}
\label{tab:dataset_counts}
\begin{tabular}{lrrrr}
\toprule
Dataset & Documents & Representations & Parquet & Tokens$^*$ \\
\midrule
Court Listener (CAP) & 6,919,296 & 6,919,272 & 6,919,272 & 16,674,704,833 \\
Court Dockets & 641,964 & 641,961 & 641,945 & 7,358,684,300 \\
.gov Websites & 3,233,136 & 3,192,174 & 3,187,571 & 22,249,301,957 \\
Electronic CFR & 262,243 & 262,243 & 262,243 & 139,308,629 \\
SEC EDGAR Filings & 74,063,501 & 30,474,244 & 44,768,118 & 975,315,045,213 \\
EU Official Journal & 1,389,632 & 1,386,410 & 1,306,253 & 52,400,358,216 \\
Federal Depository Library & 319,248 & 289,624 & 289,583 & 7,720,867,719 \\
Federal Register & 3,396,818 & 3,396,455 & 3,396,389 & 14,766,889,851 \\
GovInfo & 15,342,752 & 14,494,739 & 11,148,500 & 87,180,746,390 \\
RECAP & 16,762,471 & 14,967,921 & 14,265,800 & 65,309,281,664 \\
RECAP Documents & 1,863,733 & 1,691,658 & 1,691,655 & 5,946,067,662 \\
Regulations.gov & 1,279,349 & 1,247,138 & 1,101,913 & 9,536,541,528 \\
UK Legislation & 219,190 & 219,190 & 219,190 & 2,144,665,779 \\
US Code & 69,391 & 69,391 & 69,391 & 70,278,541 \\
USPTO Patents & 6,586,666 & 6,413,833 & 6,413,827 & 81,575,351,620 \\
\midrule
Total & 132,349,390 & 85,666,253 & 95,681,650 & 1,348,388,093,907 \\
\bottomrule
\multicolumn{5}{p{\textwidth}}{\footnotesize $^*$ Token counts are extrapolated from a 57.8M document snapshot
(representing 60.4\% of the final dataset) based on per-dataset document ratios for "useful" formats. These counts use the
\texttt{kl3m-004-128k-cased} tokenizer from \cite{bommarito2025tokenizers}. For tokenization efficiency comparisons with other
tokenizers, see Table II in that work. Additionally, true total token counts, including
low-quality PDFs and formats like XBRL, are likely on the order of 2-3x larger.} \\
\end{tabular}
\end{table}

\begin{itemize}
    \item \textbf{Varying scale:} Document volumes range from tens of millions (EDGAR, RECAP, 
USPTO) to smaller but significant collections like the US Code in Markdown format.

    \item \textbf{Processing status:} While some datasets (USPTO, Courts) have been fully 
processed through the pipeline, others remain partially processed due to resource requirements 
(e.g., OCR), prioritization, or rate of new documents (e.g., EDGAR, PACER). 

    \item \textbf{Document expansion:} The expansion of document counts between stages 
(particularly visible in EDGAR and regulatory collections) demonstrates how each originally-retrieved
document may contain multiple separate units.

    \item \textbf{Jurisdictional coverage:} The collection spans US federal and state 
jurisdictions, the UK, and the EU, providing diverse geographic and legal system representation.
\end{itemize}

Compared narrowly to other "legal" datasets, KL3M resources represent a substantial advancement. Both the 
Pile of Law \cite{henderson2022pile} and MultiLegalPile \cite{niklaus2023multilegalpile}  contain 
at least one order of magnitude less content, do not provide access to original documents or enriched 
representations, and are licensed under restrictive CC BY-NC-SA 4.0 licenses that prevent their use 
in practice.  Our dataset, though significantly less diverse than both, is also likely larger than the
original Pile \cite{gao2020pile} and on the same scale as the RedPajama resources \cite{weber2024redpajama}
once our tokenizer's relative efficiency is accounted for.\footnote{As we have only tokenized our snapshots using the
KL3M family tokenizers, we cannot conclusively confirm these comparisons to \texttt{gpt-2} or \texttt{Mistral} counts.}

\subsection{Token Statistics and Content Distribution}
\label{subsec:token_statistics}

Table \ref{tab:token_length_stats} next provides a source-specific and aggregate summary of document length characteristics.  Notable observations for these statistics include:

\begin{table}[ht]
\centering
\caption{Document Length Statistics by Dataset}
\label{tab:token_length_stats}
\begin{tabular}{lrrrrr}
\toprule
Dataset & Mean Tokens & Median Tokens & $\geq$ 8K (\%) & $\geq$ 32K (\%) & $\geq$ 100K (\%) \\
\midrule
UK Legislation & 32,675 & 2,493 & 36.1 & 21.1 & 10.1 \\
Federal Depository Library & 31,473 & 8,398 & 50.8 & 23.3 & 6.1 \\
USPTO Patents & 12,718 & 8,925 & 56.3 & 5.2 & 0.5 \\
Court Dockets & 10,388 & 2,344 & 21.1 & 3.2 & 1.1 \\
Regulations.gov & 10,354 & 2,357 & 18.4 & 5.7 & 1.7 \\
.gov Websites & 9,055 & 1,749 & 14.5 & 4.0 & 1.3 \\
RECAP & 4,053 & 977 & 11.2 & 1.2 & 0.2 \\
Federal Register & 3,865 & 1,620 & 7.2 & 1.4 & 0.3 \\
RECAP Documents & 3,560 & 1,973 & 11.3 & 0.4 & 0.0 \\
Court Listener (CAP) & 2,408 & 1,426 & 5.1 & 0.1 & 0.0 \\
US Code & 997 & 360 & 1.5 & 0.1 & 0.0 \\
Electronic CFR & 523 & 206 & 0.5 & 0.0 & 0.0 \\
\midrule
All Datasets & 6,237 & 1,855 & 17.5 & 2.4 & 0.5 \\
\bottomrule
\multicolumn{6}{p{\textwidth}}{\footnotesize Note: Statistics are based on a subset of approximately 44M computed using the \texttt{kl3m-004-128k-cased} tokenizer, excluding the EU resources as of the selected snapshot.} \\
\end{tabular}
\end{table}

\begin{itemize}
    \item \textbf{Length variation by source:} Mean document length by source ranges from tens of thousands
in some sources to just 523 tokens in the eCFR, reflecting the inherent variation in content type and 
publishing style.

    \item \textbf{Abundance of long-context material:} The collection includes substantial 
long-context training material.  Over 1 in 6 documents documents exceed 8,000 tokens, nearly 1 in 40
is at least 32,000 tokens, and over 200,000 documents exceed 100,000 tokens.  Certain collections, like the
full USPTO Granted Patents and Federal Depository Library, are particularly rich sources for coherent long-range 
generation.

    \item \textbf{Length distribution profile:} With a mean of 6,237 tokens and median of 1,855 
tokens across all sources, these materials in aggregate provide longer average document lengths than 
most Internet-sourced corpora.
\end{itemize}

\begin{figure}[htbp]
\centering
\includegraphics[width=\textwidth]{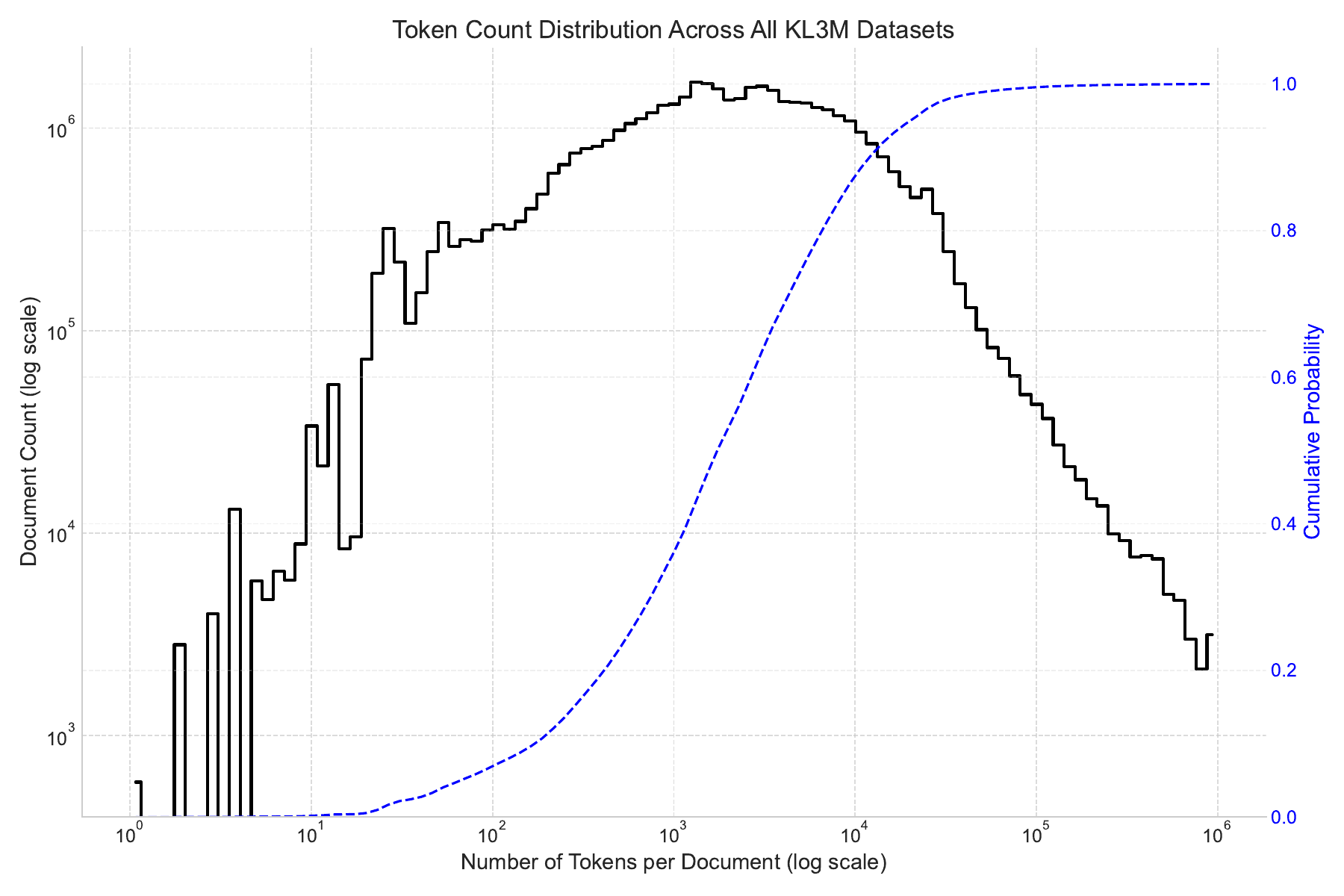}
\caption{Aggregated token count distribution across all sources. The log-log scale reveals both the 
high frequency of documents in the 1K-10K range and the substantial long tail extending beyond 100K 
tokens.}
\label{fig:aggregated_token_histogram}
\end{figure}

Figure~\ref{fig:aggregated_token_histogram} visualizes this distribution, revealing both the 
concentration of moderate-length documents and the significant long tail that makes this collection 
valuable for advanced context modeling.

\subsection{Entropy Distribution Analysis}
\label{subsec:entropy_analysis}

In addition to counting tokens, we also measured the information-theoretic properties of the collection through token-level entropy analysis. Entropy, in simple terms, measures how predictable or unpredictable language patterns are within each dataset. This reveals interesting differences within and across these sources.

\begin{figure}[htbp]
\centering
\includegraphics[width=\textwidth]{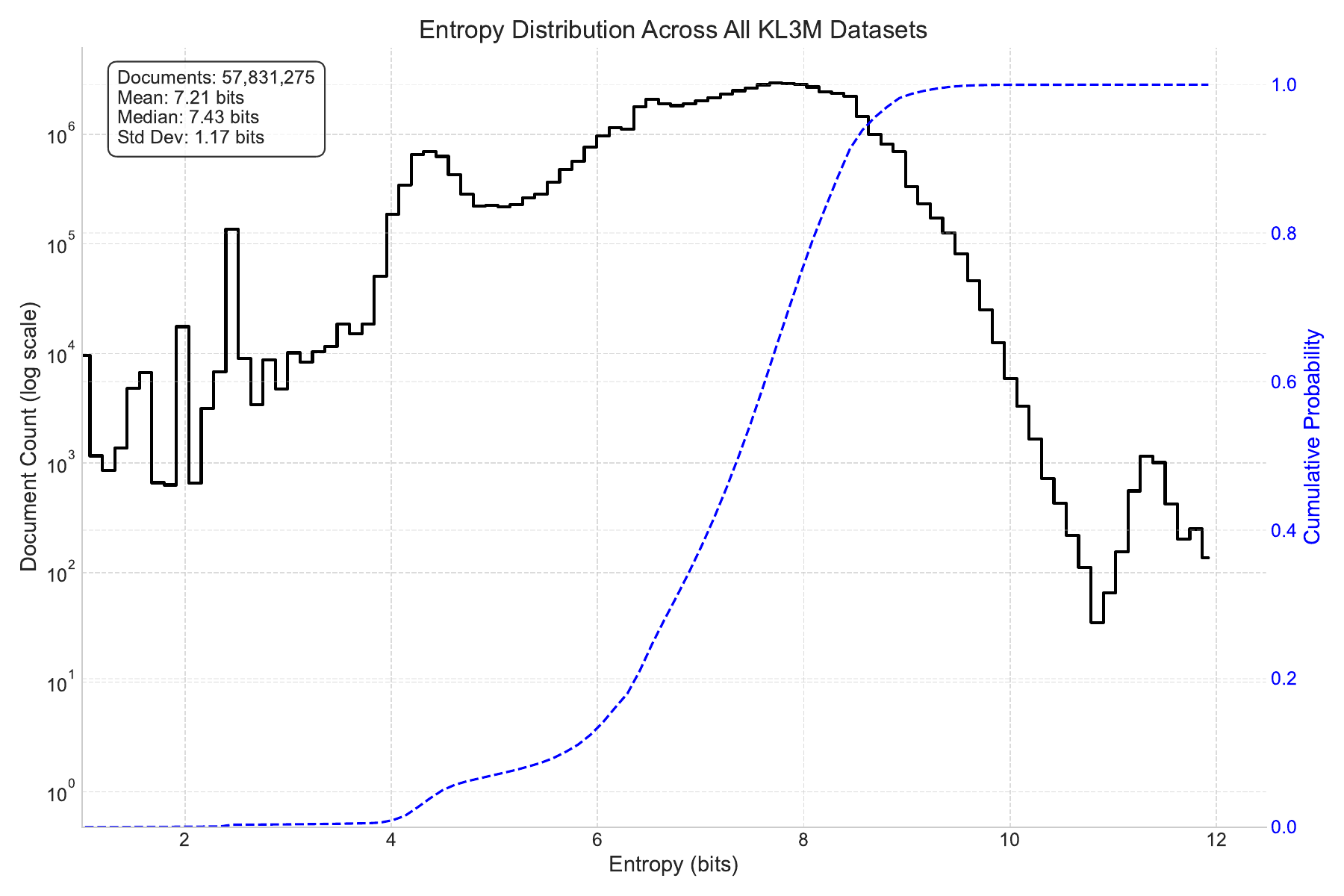}
\caption{Token-level entropy distribution across the KL3M collection. The log-scale y-axis shows 
document count, while the x-axis shows entropy in bits.}
\label{fig:entropy_histogram}
\end{figure}

Table \ref{tab:entropy_statistics} and Figure~\ref{fig:entropy_histogram} detail how entropy is distributed within and across all datasets, with most documents having values between 6-8 bits. This aligns with research by \cite{bentz2017entropy}, which estimates unigram entropy of written English to be between 7.5-10 bits depending on document length, and notes that no natural language has an estimated entropy below 6.0 bits. Table~\ref{tab:entropy_statistics} breaks this down by dataset, showing how different document types vary in their entropy characteristics.

\begin{table}[htbp]
\centering
\caption{Token Entropy Statistics by Dataset (in bits)}
\label{tab:entropy_statistics}
\begin{tabular}{lrrr}
\toprule
Dataset & Mean & Median & Std \\
\midrule
Federal Depository Library & 7.56 & 8.08 & 1.57 \\
Edgar-Agreements & 7.79 & 7.95 & 0.83 \\
EU Official Journal & 7.87 & 7.95 & 1.29 \\
USPTO Patents & 7.87 & 7.82 & 0.50 \\
RECAP Documents & 7.56 & 7.80 & 0.87 \\
Federal Register & 7.67 & 7.68 & 0.53 \\
Regulations.gov & 7.57 & 7.67 & 0.96 \\
Court Listener (CAP) & 7.27 & 7.63 & 1.18 \\
UK Legislation & 7.38 & 7.46 & 0.85 \\
SEC EDGAR Filings & 7.31 & 7.45 & 1.11 \\
.gov Websites & 7.18 & 7.32 & 1.02 \\
RECAP & 6.80 & 7.07 & 1.49 \\
GovInfo & 6.89 & 6.70 & 0.77 \\
US Code & 6.32 & 6.57 & 1.16 \\
eCFR & 6.19 & 6.29 & 0.94 \\
Dockets & 6.26 & 6.24 & 0.31 \\
\bottomrule
\end{tabular}
\end{table}

The entropy statistics reveal notable variation across dataset types:
\begin{itemize}
    \item \textbf{Federal Depository Library and EU Official Journal} exhibit the highest median 
entropy (8.08 and 7.95 bits).
    
    \item \textbf{Patents and regulatory documents} (USPTO, Federal Register) show moderately high 
entropy (7.82 and 7.68 bits).
    
    \item \textbf{Codified materials} like the US Code and eCFR have lower entropy values (6.57 and 
6.29 bits).
\end{itemize}

Entropy naturally varies across languages, and some of these sources, like the EU resources, contain multiple languages.  While we did not perform explicit language detection across all documents, language metadata is available for certain sources. Federal websites often include HTML language tags that indicate their language and locale, and European Union documents identify language through their RDF and URIs. This metadata enables filtering by language for targeted research applications. 

It is worth noting that entropy values are influenced by tokenization efficiency. The KL3M tokenizer was also optimized for legal and financial text in a limited number of languages, so documents that cover other domains and languages typically show higher entropy values. This occurs because the tokenizer must use more tokens to represent text in languages for which it wasn't specifically optimized. The variation in entropy across datasets thus partially reflects both inherent linguistic complexity and the interaction between language and tokenization approach. This reinforces the importance of specialized tokenizers when working with domain-specific or multilingual collections.

\subsection{Enterprise File Sample}
\label{subsec:enterprise_file_sample}

In addition to the three-stage training data resources, we also provide a substantial Enterprise File Sample containing nearly 500,000 document files in original "enterprise" formats like PDF, Word, Excel, and PowerPoint, that we collected from U.S. government websites. This collection, available under the \texttt{raw/dotgov/} prefix in the \texttt{data.kl3m.ai} bucket, provides researchers with realistic enterprise document formats that are generally hard to source from otherwise clean sources.  This sample can be used to enable the development and testing of document conversion and extraction tools on realistic enterprise content and metadata.

\subsection{Mid and Post-Train Resources}
\label{subsec:mid_post_train}

Beyond pre-training data, our project provides specialized datasets designed for mid-training and post-training stages of model development. These resources support supervised fine-tuning, instruction tuning, and evaluation across legal and regulatory domains. 

The current collection of resources is curated and maintained on \href{https://huggingface.co/collections/alea-institute/alea-mid-and-post-train-resources-67f5a8d7fc4b3d0ff4448aa5}{a Hugging Face collection}. The initial release includes five types of resources:

\begin{itemize}
    \item Question-answer pairs and definitions derived from documents like website FAQs and the CFR;
    \item Abstractive and extractive summarization tasks derived from document metadata and inline summaries, abstracts, conclusions, or descriptions;
    \item Classification tasks for legal and regulatory materials derived from self-reported or assigned labels;
    \item Structured document generation (e.g., patents, contracts); and
    \item Multi-turn conversations from hearings and public forums.
\end{itemize}

As noted at the beginning of this section, we intend to continue expanding the coverage, quality, and depth of these resources indefinitely.
\section{Impact and Conclusion}
\label{sec:impact_conclusion}

In this paper, we have presented the first major open milestone for this project: a comprehensive set of training data resources specifically designed to reduce the legal and ethical risks that challenge development and use of LLMs today.  The protocol and pipeline described in Sections \ref{sec:data_protocol} and \ref{sec:pipeline_implementation} directly address the critical issues identified in Section~\ref{sec:legal_problem_space}. 

The KL3M Data Project delivers the primary contributions outlined in Table~\ref{tab:contributions}, including (1) over 132 million documents and trillions of tokens from verifiably public domain or appropriately-licensed sources; (2) the complete source code to acquire and process these documents; (3) multi-stage data access with original document formats, extracted content, and pre-tokenized representations; (4) rich Dublin Core metadata with search and interactive exploration capabilities; (5) specialized mid- and post-training resources for specific legal domains or use cases; and (6) an "enterprise" document collection with nearly 500,000 original PDF and Office file formats.  These resources are all freely available to the public on S3, Hugging Face, and GitHub under permissive CC-BY and MIT terms.

The potential applications of the KL3M resources are diverse and significant. The dataset provides a comprehensive foundation for small or domain-specific model pre-training that can be supplemented with other appropriately licensed datasets, as we have already demonstrated with our SLM models like \texttt{kl3m-002-170m} or \texttt{kl3m-003-1.7b}. It also offers valuable resources for fine-tuning existing models to enhance performance on a variety of tasks, especially in the legal and financial domains.  While this dataset alone may not cover all use cases, it represents a substantial corpus that, when strategically combined with selected licensed content, could facilitate the development of high-performing LLMs that maintain legal compliance.

In contrast to practically all existing LLMs that utilize copyrighted materials obtained without consent or explicit licensing, the KL3M Data Project also establishes an alternative paradigm built on positive legal rights and consent. If artificial intelligence systems are to embody societal values and beliefs for an improved future, we believe that they must be developed \textit{within}, not outside of, our shared legal and ethical frameworks. The ongoing copyright and contract litigation documented in Section~\ref{sec:legal_problem_space} demonstrates that legal ambiguity in this domain will likely persist for the foreseeable future unless the field adopts a fundamentally different approach.

Our future research agenda extends beyond mere dataset expansion to the establishment of a federated project where researchers can leverage our infrastructure to broaden the availability of copyright-clean data across jurisdictions. While our initial focus has centered on text content in the U.S. and EU jurisdictions due to our familiarity with these legal systems, we intend to systematically incorporate content from diverse legal systems, languages, domains, and audiovisual formats so long as content can meet the test of our protocol. We have already begun to develop rigorous domain-specific evaluation benchmarks for assessing LLM performance across specialized legal tasks including statutory interpretation, case analysis, and contract review. Furthermore, we will address the challenge of attribution and temporal drift in legal and regulatory content through implementation of mechanisms to ensure models produce citation and maintain currency with evolving law.  Lastly, we intend to augment these traditional dataset collection and curation approaches with more knowledge-graph driven approaches designed to embody the original vision of the Semantic Web.

We believe that the KL3M Data Project has empirically demonstrated that large-scale, high-quality data collection can successfully operate within established legal and ethical boundaries. By building on positive legal rights and consent rather than litigation and violation of expressed preferences, we can develop AI systems that are not only useful but also legally sound and ethically grounded.

The cornerstone of our future vision is collaborative participation. We extend an invitation to researchers, legal scholars, and AI practitioners to join us in this initiative to construct a more fair, sustainable pathway to the development of this technology.  We hope you join us.

\section*{Acknowledgments}
We revised this paper with the assistance of large language models.  All errors or omissions are our own.

\bibliographystyle{plainnat}
\bibliography{main.bib} 

\end{document}